\pdfoutput=1

\documentclass[11pt]{article}

\usepackage{acl}

\usepackage{times}
\usepackage{latexsym}

\usepackage[T1]{fontenc}

\usepackage[utf8]{inputenc}

\usepackage{microtype}

\usepackage{url}
\usepackage{booktabs}
\usepackage{multirow}
\usepackage{amsmath}
\usepackage{comment}
\usepackage{graphicx}
\usepackage[whole]{bxcjkjatype}
\usepackage{amssymb}

\usepackage{xcolor}

\title{Rethinking Response Evaluation from Interlocutor's Eye\\
for Open-Domain Dialogue Systems}

\author{Yuma Tsuta\textsuperscript{1}\quad
Naoki Yoshinaga\textsuperscript{2}\quad
Shoetsu Sato\textsuperscript{2}$^*$\quad
Masashi Toyoda\textsuperscript{2} \\
 \textsuperscript{1}The University of Tokyo \\
 \textsuperscript{2}Institute of Industrial Science , The University of Tokyo \\
\texttt{\{tsuta,ynaga,shoetsu,toyoda\}@tkl.iis.u-tokyo.ac.jp}
}

\begin{document}
\maketitle
\def\thefootnote{*}\footnotetext{This affiliation is at the time of the study and differs from at the time of publication.}\def\thefootnote{\arabic{footnote}}
\begin{abstract}
Open-domain dialogue systems have started to engage in continuous conversations with humans.
Those dialogue systems are required to be adjusted to the human interlocutor and evaluated in terms of their perspective.
However, it is questionable whether the current automatic evaluation methods can approximate the interlocutor's judgments.
In this study, we analyzed and examined what features are needed in an automatic response evaluator from the interlocutor's perspective.
The first experiment on the Hazumi dataset revealed that interlocutor awareness plays a critical role in making automatic response evaluation correlate with the interlocutor's judgments.
The second experiment using massive conversations on X (formerly Twitter) confirmed that dialogue continuity prediction can train an interlocutor-aware response evaluator without human feedback while revealing the difficulty in evaluating generated responses compared to human responses.

\end{abstract}

\section{Introduction}\label{sec:introduction}
Along with the growth of open-domain dialogue systems~\citep{xu-etal-2022-beyond,xu-etal-2022-long,bae-etal-2022-keep,MeguruTakasaki:2023}, it is crucial to develop automatic methods that efficiently evaluate those systems.
The automatic evaluations usually qualify system responses for utterances sampled from human conversation logs (\S~\ref{sec:rel-work}).
Since \citet{liuHowNOTEvaluate2016} showed that automatic evaluation with a single reference response such as BLEU~\citep{papineniBleuMethodAutomatic2002,forgues2014bootstrapping} did not correlate with human judgments due to the response diversity in open-domain dialogue~\citep{sato-etal-2017-modeling,tsuta-etal-2020-ubleu}, unsupervised reference-free methods and supervised methods that mimic human judgments have become popular~\citep{yeh-etal-2021-comprehensive}.
However, these studies evaluate their methods in terms of correlation with judges by third-party annotators (\textit{outsiders}), not partaking in the dialogue.

\begin{figure}[t]
\centering
\includegraphics[width=0.90\linewidth]{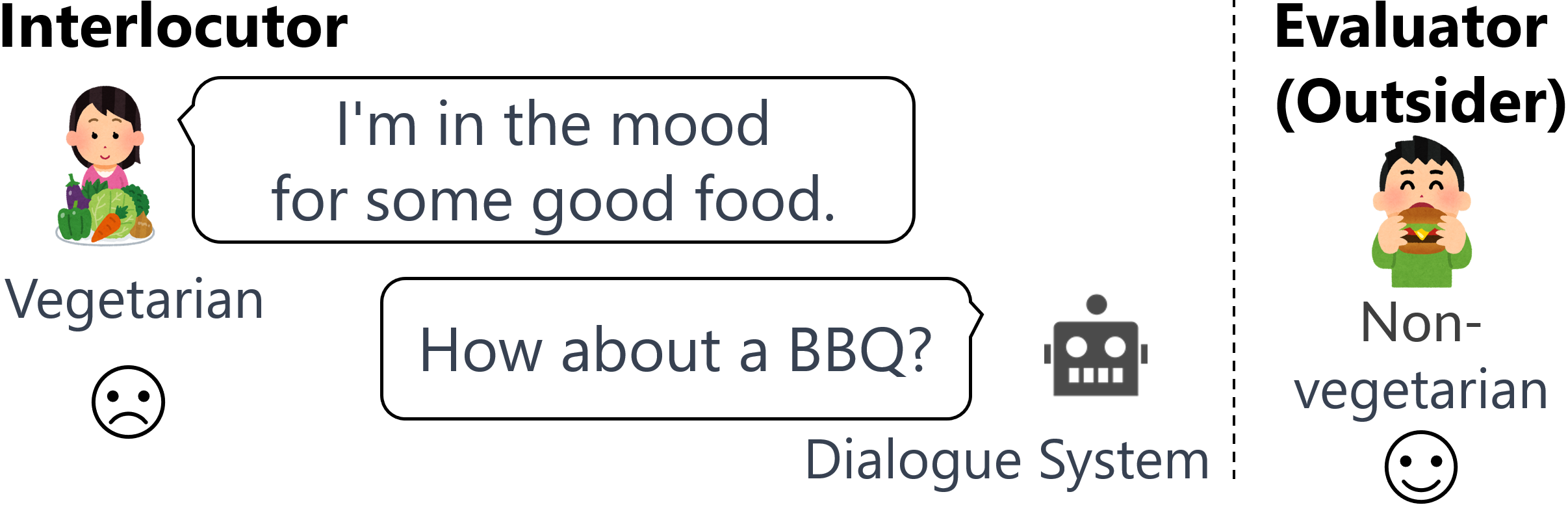}
\caption{A discrepancy between interlocutor and outsider evaluations for open-domain dialogue systems.}
\label{fig:rq}
\end{figure}

Do the existing methods correctly evaluate the dialogue systems?
As illustrated in Figure~\ref{fig:rq}, the interlocutor and evaluators may prefer different yet valid responses.
Although \citet{ghazarian-etal-2022-wrong} experimentally confirmed a poor correlation with outsider and interlocutor evaluations in terms of appropriateness, they remain focused on outsider evaluations.
This study focuses on interlocutor evaluations to enable an automatic evaluation from the interlocutor's perspective.
In the experiment, we concentrate on validating our ideas in terms of engagement.
Because this metric is more subjective and varies across people.

In this study, for estimating the interlocutor's evaluations, we first analyze the effectiveness of 
personalizing the evaluation model to the target interlocutor.
This is inspired by research on response generation~\citep{li-etal-2016-persona,xu-etal-2022-beyond}, as it has been reported to be important to adjust (personalize) utterances to the interlocutor.
For this analysis, we used the Hazumi dataset~\citep{9597447}, and confirmed that, even when we train a supervised evaluator to mimic interlocutor scores, it cannot accurately predict their scores without making it aware of the target interlocutor.

\begin{figure}[t]
\centering
\includegraphics[width=0.7\linewidth]{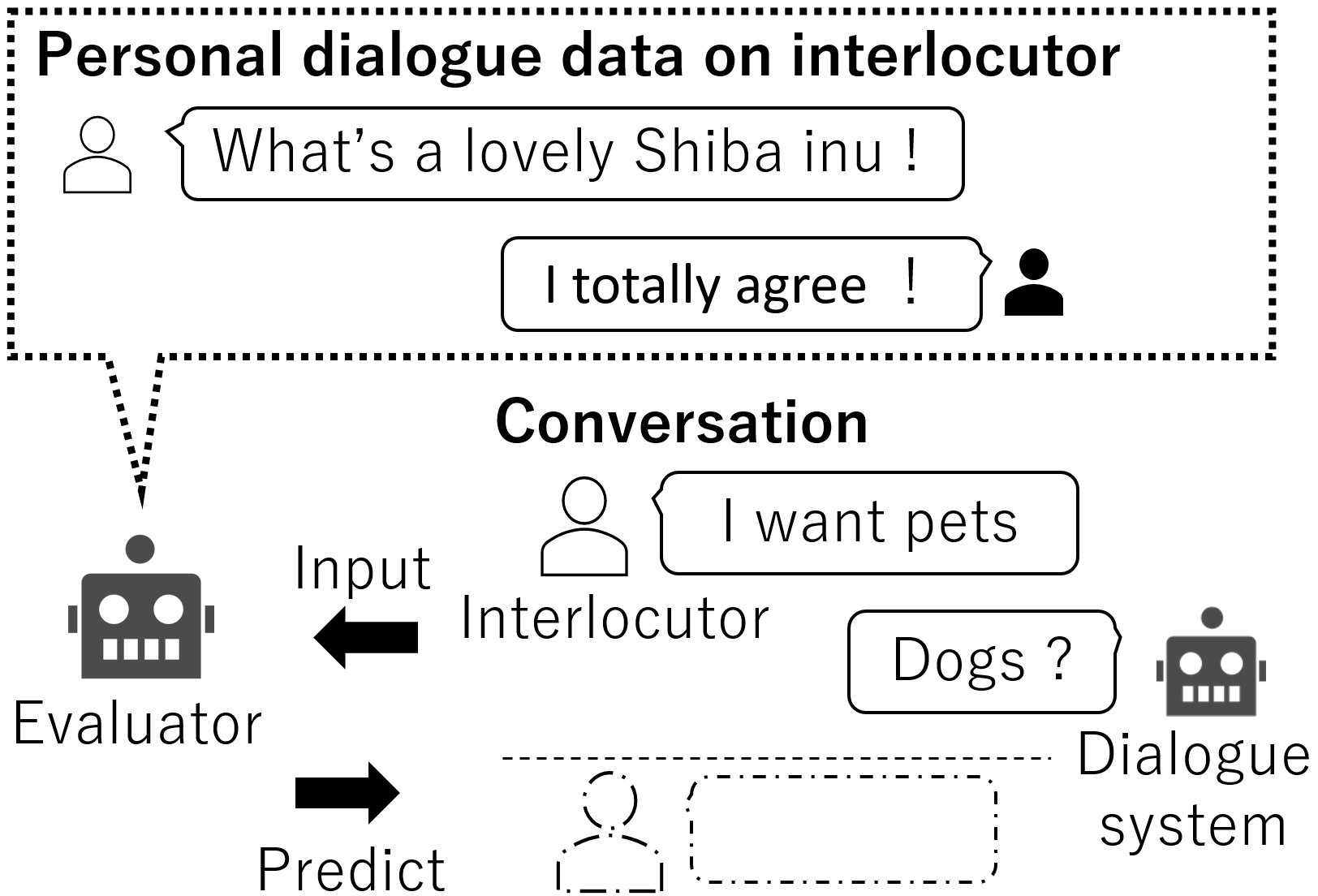}
\caption{Automatic response evaluation via dialogue continuity prediction from the interlocutor's perspective.}
\label{fig:dcp-task}
\end{figure}

Motivated by the lessons learned from the above experiments, we then explore automatic response evaluation from the interlocutor's eye (\S~\ref{sec:proposal}). 
To reduce the cost of annotation, we utilize a dialogue continuity prediction (DCP) task to train an interlocutor-aware evaluator (Figure~\ref{fig:dcp-task}).
This task of estimating whether the target speaker will continue speaking or not can take advantage of labels (conversation stop signals) that are naturally annotated by the interlocutor in the conversation log.
Experimental results on a conversation log on X (formerly Twitter) confirmed that the interlocutor-aware evaluator can be learned through the DCP task without human feedback while revealing the challenge of evaluating the system responses.

\section{Related work}\label{sec:rel-work}

\paragraph{Automatic evaluation of dialogue systems}
To efficiently develop open-domain dialogue systems, researchers have sought evaluation methods that correlate with human evaluations.
Since \citet{liuHowNOTEvaluate2016} showed that reference-based metrics~\citep{papineniBleuMethodAutomatic2002,forgues2014bootstrapping} using single reference responses do not correlate with human judgments, some studies use multiple reference responses~\citep{galley-etal-2015-deltableu,guptaInvestigatingEvaluationOpenDomain2019,tsuta-etal-2020-ubleu}, while others train models by referring to human judgments~\citep{lowe-etal-2017-towards,Ghazarian_Weischedel_Galstyan_Peng_2020} or other cues indicating valid responses~\citep{taoRUBERUnsupervisedMethod2018,ghazarianBetterAutomaticEvaluation2019,gao-etal-2020-dialogue,mehri-eskenazi-2020-usr,xu-etal-2022-endex,ghazarian-etal-2022-wrong}.
Recent studies~\citep{mehri-eskenazi-2020-unsupervised,Zhang2021} rely on language comprehension skills of pre-trained language models such as BERT~\citep{devlin-etal-2019-bert} and GPT-2~\citep{radfordLanguageModelsAre}.
There are evaluation tasks from the other perspective such as dialogue breakdown detection~\cite{higashinaka-etal-2016-dialogue}.
The above studies were, however, developed to follow outsider evaluations and do not assume evaluation from the interlocutor's eye, even though some recent dialogue systems are being adapted to interlocutors in long-term conversations~\citep{xu-etal-2022-beyond,xu-etal-2022-long,bae-etal-2022-keep,MeguruTakasaki:2023}.

A few studies have elucidated the relationship between user personality and the performance of dialogue systems from a psychological perspective~\citep{guo-etal-2021-influence,10.1007/978-981-19-5538-9_12}.
These studies suggest the importance of the interlocutor's traits in evaluating dialogue systems.

\paragraph{User-oriented NLP tasks}
There are several user-oriented (or personalized) NLP tasks in which users prefer different outputs and hence the systems are expected to be adjusted to match user preferences, 
including hashtag recommendations on social networking sites~\citep{10.1007/978-3-642-35386-4_25} and website recommendations~\citep{MISHRA20151}.
Similarly, for text generation tasks in which models have become able to generate decent outputs, researchers are starting to adapt the models to reflect individual preferences;
examples of such tasks include summarization~\citep{DIAZ20071715}, machine translation~\citep{mirkin-meunier-2015-personalized}, text simplification~\citep{bingel-etal-2018-lexi}, and dialogue systems~\citep{liu-etal-2020-impress,cho-etal-2022-personalized}.
To evaluate these systems, they need human judgments by the system users, which low reproducibility prevents us from efficiently developing the systems.

\section{What is important to predict interlocutor evaluations?}\label{sec:prelim}
To analyze what features are important for predicting interlocutor scores, we train score prediction models with several settings and compare their performances.
Specifically, we analyzed the effect of reference scores (\textit{e.g.}, interlocutor or outsider scores) and interlocutor-aware personalization on the evaluation models.
Although \citet{ghazarian-etal-2022-wrong} confirmed a low correlation between interlocutor and outsider evaluations, we further confirmed that outsider evaluations do not help predict interlocutor scores.
For this analysis, we used the Hazumi dataset~\citep{9597447}, which is an open-domain conversation in the form of the Wizard of Oz experiment.

\subsection{Hazumi dialogue datasets}\label{sub-sec:prelim-data} 
For this analysis, we need a dataset that contains interlocutor and outsider scores to train and test models, and we utilize Hazumi1902 and Hazumi1911 subsets from the Hazumi dataset\footnote{\url{https://www.nii.ac.jp/dsc/idr/en/rdata/Hazumi/}}.
This dataset is an open-domain conversation in which ``Wizard'' behaves like a dialogue system and ``Participant'' speaks as the user.
These subsets only contain an interlocutor's (\textit{e.g.}, Participant's) and five outsiders' scores for each utterance by the Wizard.
The participants and five outsiders rated the Wizard's utterances on a scale of 1 (feeling negative) to 7 (feeling positive) on the basis of user impressions.\footnote{The annotation guidelines for user impressions define keywords and keyphrases such as ``wants to keep talking'' and ``satisfied'' as positive impressions, and ``doesn't want to keep talking'', ``frustrated'' and ``confused'' as negative impressions.} 
The direction of the guideline is similar to the engagement metric in \citet{Ghazarian_Weischedel_Galstyan_Peng_2020} and the annotation on the experiment in \S~\ref{sec:experiment} in terms of the willingness of dialogue continuity.

In what follows, we preprocess the dataset so that the exchanges, a pair of utterances by a Wizard and the Participant, consist of no empty utterance.
After these preprocessing steps, we obtained 5301 exchanges from 60 dialogues.
The detailed statistics are shown in Table~\ref{tab:static-hazumi}.
We split each conversation into 8:1:1 size chunks according to the flow of the conversation (and recombined) to train, validate, and test the prediction models.

\begin{table}[t]
\centering
\tabcolsep3.2pt
\small
\begin{tabular}{lcc}\toprule
    Dataset & Hazumi1902  & Hazumi1911 \\\midrule
    \# dialogues (participants) & 30 & 30 \\
    Total count of exchanges
    & 2477  
    & 2824  \\
    Utterance length (Wizard) & 22.7 & 20.8  \\
    Utterance length (Participant) & 22.2 & 25.1 \\  \bottomrule
\end{tabular}
\caption{\label{tab:static-hazumi}
Statistics of the subsets of the Hazumi datasets after preprocessing (\S~\ref{sub-sec:prelim-data}).
Utterance length refers to the average number of characters in an utterance.}
\end{table}

\subsection{Analyze the effective cues in interlocutor score prediction}\label{sub-sec:prelim-pred}
We train evaluators using various cues to identify the interlocutor scores and clarify the requisite for automatic interlocutor evaluation.
In this task, the models predict the interlocutor score to an utterance by Wizard.
We feed Wizard's utterance and the longest contexts possible to the model, adding a special speaker token (\texttt{[Wizard]} or \texttt{[Participant]}) to distinguish who speaks utterances before the corresponding utterances.

\paragraph{Models}
We compared four evaluator models based on BERT~\citep{devlin-etal-2019-bert} for ablation. 
The differences between these models are i) whether to use interlocutor scores or (the averaged) outsider scores as the reference in training and ii) whether to use a speaker token specific to the target participant or the generic participant token preceding each utterance. 
The distinguished participant token is meant to adjust the evaluator to individual interlocutors, inspired by the speaker token introduced by \citet{li-etal-2016-persona} to model speakers in response generation.

\paragraph{Settings}
We fine-tuned each model from pre-trained Japanese BERT\footnote{\url{https://huggingface.co/cl-tohoku/bert-base-japanese-v2}\label{url:bert}} for 10 epochs with the mean squared error loss.
Other settings for the model were as follows: learning rate was $3e-5$ and optimized with AdamW~\citep{kingmaAdamMethodStochastic2015}, and batch size was 64. 
We stored the model after each epoch and adopted the model that achieved the lowest loss for the validation data for testing.

\begin{table}[t]
\centering
\tabcolsep5pt
\small
\begin{tabular}{ccr}\toprule
    Training score & Target awareness  & Pearson's $r$ \\\midrule
    Outsider & & $0.141$  \\
    Outsider & \checkmark & $0.142$  \\
    Interlocutor &  & $0.166$ \\
    Interlocutor & \checkmark & $\textbf{0.496}$ \\
    \bottomrule
\end{tabular}
\caption{Results on interlocutor score prediction with ablation of training score and target speaker awareness.\label{tab:result-pelim-pred}}
\end{table}

\paragraph{Results}
Table~\ref{tab:result-pelim-pred} shows the correlations between model predictions and interlocutor actual scores.
When a model is trained to predict the averaged outsider score, the evaluator showed a very low correlation of about 0.14.
This confirms that outsider scores are useless in predicting interlocutor scores.
Meanwhile, the model exhibits a much higher correlation when trained to predict interlocutor scores only with the awareness of target interlocutors; otherwise, the model shows only a slight improvement over the model learned by the averaged outsider scores.
These results suggest that automatic interlocutor evaluation requires us to not only take the interlocutors' view (here, scores) into account but also to be aware of the target interlocutor.

\section{Towards Automatic Response Evaluation from Interlocutor's Eye}\label{sec:proposal}
From the result in \S~\ref{sec:prelim}, we confirmed that accurate interlocutor score prediction requires personalizing the evaluator to the target interlocutor as well as referring to interlocutor scores.
In practice, however, collecting interlocutor scores and creating conversations for the annotation are costly.

Therefore, focusing on evaluating responses in terms of engagement, we propose an alternative method to train an interlocutor-aware response evaluator via a dialogue continuity prediction task, assuming that utterances replied to by the interlocutors are more engaging than utterances without a response.
The task is to predict whether there will be a response to an utterance in dialogue.\footnote{Although \citet{ghazarian-etal-2022-wrong} has also utilized the dialogue continuity prediction task for evaluating dialogue systems.
They requested the interlocutors to explicitly expose to spoken dialogue systems whether or not to stop conversations, whereas we collected this label as an implicit signal from no response in human conversation logs.}

\subsection{Interlocutor Evaluation via Personalized Dialogue Continuity Prediction (DCP)}\label{sub-sec:dcp}
We train an automatic response evaluator via the dialogue continuity prediction task (Figure~\ref{fig:dcp-task}).
The task settings are as follows. 
The task input is a conversation containing $N$ utterances $U = \{u_0,u_1,... ,u_{N-1}\}$ made by two speakers $s_i$ and $s_j$ ($u_{N-1}$ is made by $s_j$).
The model output is assumed as the probability of whether the next response $u_N$ is made by $s_i$, $P({u_N} = \textit{exists} ~|~ U, s_i)$.

\paragraph{How to consider the interlocutor in a model?}
As we have observed in \S~\ref{sub-sec:prelim-pred}, it is crucial to personalize a response evaluator to the target interlocutor to estimate human judgments given by the interlocutors.
Inspired by existing studies on personalizing open-domain dialogue systems~\citep{li-etal-2016-persona,zhang-etal-2018-personalizing}, we consider two methods for the evaluator to take the interlocutor into account.
The first method leverages a speaker token specific to the target interlocutor, which has been used in the experiments in \S~\ref{sub-sec:prelim-pred}, whereas the second method refers to a user profile of the interlocutor.
When we train a speaker token specific to the target interlocutor, we follow the procedure described in \S~\ref{sub-sec:prelim-pred}.
When using the profile, we input the profile text that accompanies the evaluation datasets (\S~\ref{subsec:corpus}) at the beginning of the model inputs.
We also consider the combination of a speaker-specific token and profile.
In summary, we use three methods to model the interlocutor: using a speaker-specific token, using the profile, and using both methods simultaneously.

\subsection{Experimental Setup}\label{sec:experiment}
To investigate the effectiveness of our interlocutor-aware evaluators, we conduct experiments focusing on two metrics: 1) 
\textbf{accuracy of the dialogue continuation task} and 2) \textbf{correlation with manually-annotated engagement scores}.

\begin{table}[t]
\centering
\small
\tabcolsep2.7pt
\begin{tabular}{lrrr}\toprule
    Data Type  & \multicolumn{1}{c}{Train} & \multicolumn{1}{c}{Dev.} & \multicolumn{1}{c}{Test} \\\midrule
    Avg. turns in dialogue  & 3.4 & 3.4 & 3.3  \\
    Avg. char. size in turn      & 31.0 & 31.1 & 30.6  \\
    Avg. char. size in dialogue  & 106.5 & 106.8 & 101.2  \\
    Replied response size & 1,779,895 & 100,899 & 1,088,970\\
    No replied response size & 1,244,530 & 70,135 & 832,377\\\bottomrule
\end{tabular}
\caption{Statistics of the X dialogue datasets.}\label{tab:data_detail}
\end{table}

\paragraph{X (formerly Twitter) dialogue dataset}\label{subsec:corpus}
We conducted the experiments using conversation logs on X\@.
We can identify the author of a post, and handle a variety of users.
We developed a Japanese dialogue dataset between two users using the API\footnote{\url{https://developer.twitter.com/en/docs/twitter-api}}\@.
During the construction, we excluded posts that could be noisy, such as repetitive posts by bots, and preprocessed posts referencing studies using dialogues on Twitter~\citep{li-etal-2016-persona,tsuta-etal-2020-ubleu}.
In addition, we used only the conversations where all responses were made within 30 minutes because response rates tend to decrease over time~\citep{gao-etal-2020-dialogue}.
We expect these processes to make conversations more engaging, coherent, and less interrupted by others.

We randomly select 10,000 users who have had at least 30 conversations between January 2017 and March 2018.
We use up to 400 conversations per user and their profile text to train the evaluator models.
The profiles are collected with a field of the API (\texttt{user.fields=description}) and the average character size is $75.0$.
We used conversations of these users between March and December 2018 as test data. 
Because the intermediate reply is a positive sample and the last reply is a negative sample in the DCP task, several samples are collected from one conversation.

For the second experiment, we need conversations between a human (interlocutor) and a dialogue system, and the interlocutor's engagement score of willingness to reply to the system responses.
Thus, we collected personal conversations on X by two members of our research group (a co-author and a graduate student) using the above same process.
The dialogue data was added to the above dataset for (19, 6, and 10) and (165, 43, and 27) conversations as training, validation, and test data, respectively.
Table~\ref{tab:data_detail} shows the statistics of the entire dataset.\footnote{The post IDs for datasets other than annotator conversations can be available on \url{http://www.tkl.iis.u-tokyo.ac.jp/\%7Etsuta/aacl-srw}.}

\paragraph{Dialog systems}\label{sec:dialog-sysytems}
To obtain system responses for human annotation, we employed seven dialogue models with two types of base architectures, Transformer encoder-decoder and decoder-only Transformer (GPT-2).
As the encoder-decoder model, we used three publicly available dialogue systems that were trained with different datasets~\citep{sugiyama2021empirical}.\footnote{\url{https://github.com/nttcslab/japanese-dialog-transformers}}
As GPT-2, We fine-tuned a pre-trained GPT-2\footnote{\url{https://huggingface.co/rinna}\label{rinna}} (medium) with our dataset (\S~\ref{subsec:corpus}).
We prepared four variations of fine-tuned GPT-2 to obtain dialogue systems with diverse conversation abilities. 
The two options are i) whether to re-initialize the model's parameters before fine-tuning and ii) whether to personalize the system to the interlocutor using a speaker token~\citep{li-etal-2016-persona}.    

\paragraph{Annotation with interlocutor judgments}
To obtain manually annotated scores to responses for the second experiment,
we asked the two annotators (same as the two interlocutors) to score seven responses generated by the above dialogue systems and one ground-truth response in the test data on a scale of 0 to 100, referring to \citet{ji-etal-2022-achieving}. 
0 means that the annotator never responds to the last utterance of the conversation, and 100 means the opposite.
We compensated the annotators at the rate of 1,050 JPY per hour.

\paragraph{Evaluator and baselines}\label{sec:setting}
We compare the following evaluation models.
Because we also evaluate actual human responses, we use reference-free evaluation models that are easily available in our Japanese corpus as baseline models:
BERT-NSP~\citep{devlin-etal-2019-bert}\footnote{We employed the next sequence prediction task as an automatic evaluation model as in other studies~\citep{mehri-eskenazi-2020-usr,phy-etal-2020-deconstruct}.}, BERT-RUBER~\citep{ghazarianBetterAutomaticEvaluation2019}, FED~\citep{mehri-eskenazi-2020-unsupervised}\footnote{We translated the follow-up utterances to evaluate system responses in terms of engagement.} and Deep-AM-FM~\citep{Zhang2021}.
We also adopt the simple baseline model that always outputs the majority class label (i.e., whether or not to reply) based on the training data.
We prepared two types of majorities: all users' majority (\textbf{Global majority}) and each interlocutor's majority (\textbf{Private majority}).\footnote{Note that we use this model only in the first experiment because the second experiment evaluates models by Pearson correlation, but this model can output either 0 or 1.}

For the baseline models, we adopted a pre-trained BERT\footref{url:bert} for BERT-* and Deep-AM, and GPT-2\footref{rinna} (small) for FED and Deep-FM.
We trained models again for domain adaptation for FED and Deep-AM-FM, and additionally fine-tuned them for BERT-* using training data.\footnote{We created their negative samples with the same amount of positive samples in the training data by randomly combining a dialogue context utterances and a reply.}
For our evaluator models, we trained BERT through the DCP task without the target user awareness (BERT-DCP) and with the personalization using user-specific token (+ user token), profile text (+ profile), or both of them (+ both).
The hyperparameters of all models were as follows: learning rate as $3e-5$, batch size as $64$, and number of epochs as $5$.
We used AdamW~\citep{kingmaAdamMethodStochastic2015} as the optimizer and cross-entropy loss as the loss function. 
All model parameters trained on our dataset, including the annotator's conversation for the second experiment, are shared across all experiments.

\begin{table}[t]
\centering
\tabcolsep 7pt
\small
\begin{tabular}{lcc}\toprule
     Evaluator & Accuracy & Macro-$F_1$  \\\midrule
    Global majority & $0.564$ & $0.361$ \\
    Private majority & $0.683$ & $0.659$ \\ \midrule
    BERT-NSP & $0.548$ & $0.444$ \\ 
    BERT-RUBER & $0.541$ & $0.488$ \\ 
    Deep-AM & $0.507$ & $0.495$ \\ 
    Deep-FM  & $0.543$ & $0.533$ \\ 
    Deep-AM-FM  & $0.541$ & $0.531$ \\ 
    FED & $0.460$ & $0.446$ \\ \midrule
    BERT-DCP & $0.668$ & $0.653$ \\ 
     + user token & $\textbf{0.751}$ & $\textbf{0.744}$ \\
     + profile & $0.746$ & $0.738$ \\
     + both & $\textbf{0.751}$ & $\textbf{0.744}$ \\
    \bottomrule
\end{tabular}
\caption{Binary classification result of dialogue continuity prediction task on X dialogue dataset.\label{tab:result}}
\end{table}

\subsection{Results}\label{sub-sec:result}
Table~\ref{tab:result} lists the results of binary classification on the dialogue continuity prediction task in terms of accuracy and macro-F$_1$ to correct label bias.
To compare the baseline model which does not output probabilities (Deep-AM-FM, FED), the model output is binarized using a threshold based on the whole user response ratio in the validation data.
Unsurprisingly, BERT-DCP fine-tuned through DCP task performed better than the baselines.
The evaluator can work with the DCP task by considering the interlocutor and get better results than \textbf{Private majority}.
We also observed that using a unique speaker token for each interlocutor was a more effective way of taking interlocutors into account.

\begin{table}[t]
\centering
\tabcolsep 5.5pt
\small
\begin{tabular}{lrrrr}\toprule
    Evaluator & \multicolumn{2}{c}{Annotator 1} & \multicolumn{2}{c}{Annotator 2} \\ 
    \cmidrule(lr){2-3}
    \cmidrule(lr){4-5}
    & Human & System &  Human & System \\\midrule
    BERT-NSP & $0.477$ & $\textbf{0.416}$ & $-0.028$ & $\textbf{0.337}$ \\ 
    BERT-RUBER & $0.285$ & $0.243$ & $-0.134$ & $0.210$ \\ 
    Deep-AM & $0.564$ & $0.293$ & $0.015$ & $0.242$\\ 
    Deep-FM  & $0.499$ & $0.031$ & $-0.011$ &  $-0.070$ \\ 
    Deep-AM-FM  & $0.528$ & $0.074$ & $-0.010$ & $-0.055$ \\ 
    FED  & $0.210$ & $0.051$ & $0.040$ & $-0.070$ \\\midrule
    BERT-DCP & $0.646$ & $0.401$  & 0.578 & $0.156$ \\
     + user token & $0.720$ & $0.364$  & $\textbf{0.582}$ & $0.072$ \\
     + profile & $\textbf{0.754}$ & $0.369$  & $0.543$ & $0.077$ \\
     + both & $0.727$ & $0.367$ & $0.527$ & $0.078$ \\
    \bottomrule
\end{tabular}
\caption{Correlation with human judgment for responses by humans (Human) and dialogue systems (System).}\label{tab:result-second}
\end{table}

Table~\ref{tab:result-second} lists the results of Pearson's $r$ correlation between each evaluator's outputs (probabilities) and interlocutor scores.
Our evaluators, BERT-DCP, have higher correlations with fluent human responses than baseline evaluators, and the improvement of performance by considering personality can be confirmed.
This result confirms the usefulness of the DCP task for predicting interlocutor evaluations.
In contrast, the BERT-NSP has the highest correlation in the system response, and all BERT-NSPs are worse than the performance in the human response.
This may be because the DCP task is trained based on real conversations and is therefore vulnerable to non-fluent and inappropriate responses by the system.
A similar tendency of lower correlation with human judgments for system responses than those for human responses has been reported for the other evaluation models on engatement~\citep{Ghazarian_Weischedel_Galstyan_Peng_2020,gao-etal-2020-dialogue}.

Because our interlocutor-aware evaluators correlate well with interlocutors' judgments of human responses, 
our method will be more useful as dialogue systems converse more naturally like humans.
However, we still need to improve the evaluator so that it is capable of evaluating dialogue systems in the future.

\begin{figure}[t]
\centering
\includegraphics[width=\linewidth]{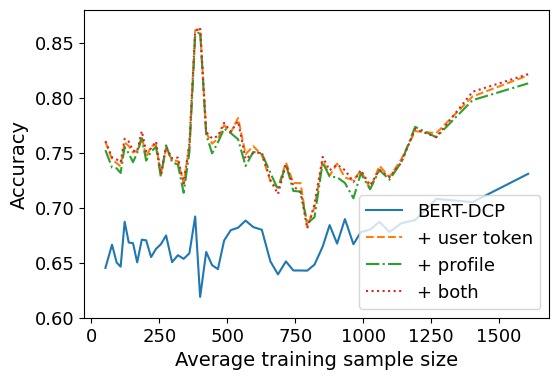}
\caption{Result of dialogue continuity prediction task per user group split according to training sample size.}
\label{tab:result-per-size}
\end{figure}

\subsection{Discussion}
The performance of our interlocutor-aware evaluator will be affected by the size of the conversation logs given by the target interlocutor.
For example, the performance could be poor for users who have a few conversations in the training data.
To investigate the relationship between the training sample size for the target interlocutor and the performance of our models, we divide the test dataset into three user groups so that the training sample size for each group is as equal as possible.
As a result, the average sample size for each group was approximately 60,000, and the smallest group had an average of 51.1 samples.
Table~\ref{tab:result-per-size} shows the result on each user group in the test dataset.
We confirmed that, with the exception of a peak around 400 samples, the accuracy changed only slightly below 1200 samples, improved above 1200 samples, and overall, the personalized models outperformed the BERT-DCP.

\section{Conclusions}
This study first explored the effect of interlocutor awareness on predicting interlocutor evaluations and then examined an automatic response evaluation method grounded in the perspective of the interlocutor.
In the first experiment using the Hazumi dataset, we confirmed interlocutor score prediction requires personalization for interlocutor awareness as well as interlocutor scores.
In the second experiment using conversations on X (formerly Twitter), we confirmed that dialogue continuity prediction is effective in training our interlocutor-aware automatic evaluator and the evaluator correlates with the actual interlocutor evaluations on human responses, while the improvement of the evaluation for the system responses is future work.

We plan to leverage recent response generation methods in long-term conversations~\citep{xu-etal-2022-beyond,xu-etal-2022-long,bae-etal-2022-keep,MeguruTakasaki:2023} to personalize our evaluator.

\section*{Acknowledgment}
This work was partially supported by the special fund of Institute of Industrial Science, The University of Tokyo, by JSPS KAKENHI Grant Number JP21H03494, and by JST, CREST Grant Number JPMJCR19A4, Japan.

\section*{Limitations}\label{sec:limit}
Although this study illuminates the demand for evaluation from the perspective of the interlocutor, we only confirmed evaluation in terms of engagement.
As existing studies on evaluation for open-domain dialogue systems are conducted in a variety of metrics such as understandability and informativeness, etc,~\cite{finch-etal-2023-dont}, interlocutor-aware evaluation in the other evaluation metrics needs to be investigated.

To realize the study for a variety of metrics, a dataset with sufficient size of conversations and annotations is needed.
In this study, we conducted experiments with two annotators to compare the automatic evaluators, but it is desirable to be annotated by a variety of people.
Therefore, it is necessary to overcome the difficulties of the cost of constructing a dataset that includes conversations with multiple dialogue systems and annotations by the speakers, as well as the privacy issues related to dataset publication to reproduce experiments.

\bibliography{aacl-srw2023}
\bibliographystyle{acl_natbib}

\appendix

\end{document}